\PassOptionsToPackage{table}{xcolor}
\documentclass{bmvc2k}

\usepackage{algorithm}
\usepackage{algorithmic}
\usepackage{booktabs}
\usepackage{multirow}
\usepackage{xcolor}
\usepackage{amsmath}
\usepackage{amssymb}
\usepackage{amsthm}

\makeatletter

\@ifundefined{theorem}
  {\newtheorem{theorem}{Theorem}}
  {}

\@ifundefined{lemma}
  {\newtheorem{lemma}{Lemma}}
  {}

\@ifundefined{proposition}
  {}
  {}

\@ifundefined{assumption}
  {\newtheorem{assumption}{Assumption}}
  {}

\@ifundefined{definition}
  {}
  {}

\@ifundefined{corollary}
  {}
  {}

\makeatother

\title{
Triple Expert Learning from Noisy Labels for
Semi-Supervised Vision Foundation Model Adaptation
}

\addauthor{Xuanyu Liu$^{*}$}
{xuanyu.liu@warwick.ac.uk}
{1}

\addauthor{Zheng Fang$^{*}$}
{Zheng.Fang.6@warwick.ac.uk}
{1}

\addauthor{Hongyang He$^{\dagger}$}
{Hongyang.He@warwick.ac.uk}
{1,2}

\addauthor{Yundi Hong}
{Yundi.Hong@warwick.ac.uk}
{1}

\addauthor{Daizong Liu}
{daizongliu@whu.edu.cn}
{3}

\addinstitution{
Department of Computer Science\\
The University of Warwick\\
Coventry, UK
}

\addinstitution{
Manifolda.AI\\
London, UK
}

\addinstitution{
Institute for Math \& AI\\
Wuhan University\\
Wuhan, China
}

\runninghead
{Liu, Fang, He, Hong, and Liu}
{Triple Expert Learning from Noisy Labels}

\begin{document}

\maketitle

\begingroup
\renewcommand{\thefootnote}{\fnsymbol{footnote}}
\footnotetext[1]{Xuanyu Liu and Zheng Fang contributed equally.}
\footnotetext[2]{Corresponding author: Hongyang He.}
\endgroup


\maketitle

\begin{abstract}
Semi-supervised adaptation of vision foundation models (VFMs) commonly freezes the pretrained backbone and updates lightweight modules such as LoRA. 
However, pseudo-labels have mixed reliability, and a single LoRA adapter must absorb reliable, ambiguous, and noisy gradients in the same low-rank space. 
This can make VFM adaptation sensitive to pseudo-label noise. 
We propose \textbf{TriNoL}, a \textbf{Tri}ple-expert learning framework from \textbf{No}isy \textbf{L}abels for semi-supervised VFM adaptation. 
TriNoL routes unlabeled samples into three confidence regions and assigns them to three LoRA experts: a Positive Expert for high-confidence pseudo-labels, an Alignment Expert for medium-confidence ambiguous samples, and a Negative Expert for low-confidence noisy samples. 
The VFM backbone remains frozen, and only the LoRA experts and classifier head are updated. 
By separating different pseudo-label reliability regions into specialized adaptation paths, TriNoL improves robustness to noisy supervision while keeping the training cost low.
\end{abstract}

\begin{figure*}[htp]
    \centering
    \includegraphics[width=\linewidth]{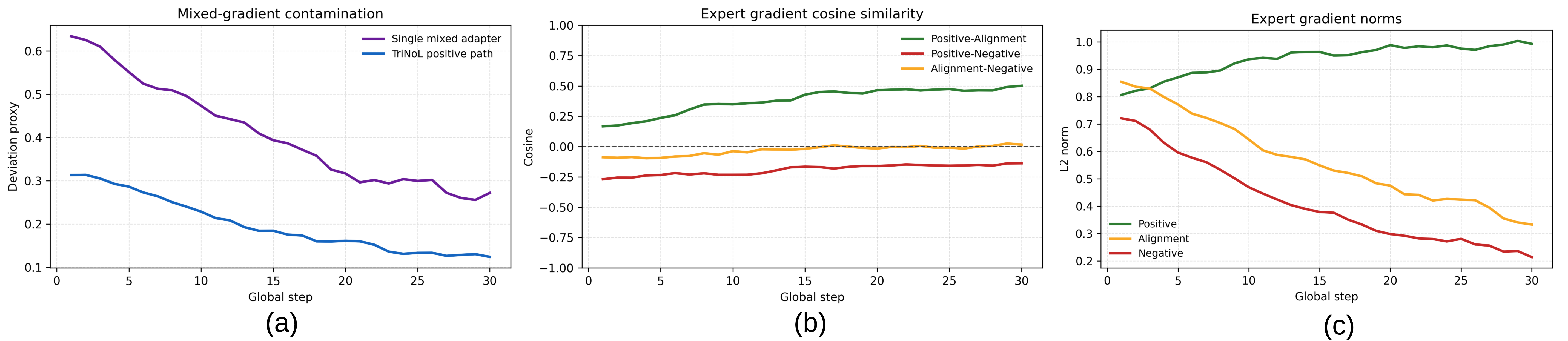}
    \vspace{-5mm}
    \caption{
    \textbf{Pseudo-labels exhibit mixed reliability across confidence regions.}
    }
    \label{fig:motivation_mixture}
\end{figure*}

\vspace{-5mm}

\section{Introduction}
\label{sec:intro}
\vspace{-3mm}

Semi-supervised learning (SSL) aims to improve visual recognition by using a small labeled set and a large unlabeled set~\cite{A2,A3,A4}. 
Modern SSL methods usually rely on pseudo-labeling and consistency regularization~\cite{A1,A2,A3,A4}. 
A model first predicts labels for weakly augmented unlabeled samples, and then uses these pseudo-labels to supervise strongly augmented views~\cite{A4}. 
This simple design has shown strong performance in many classification benchmarks. 
However, its effectiveness depends heavily on pseudo-label quality. 
When pseudo-labels are correct, they provide useful extra supervision. 
When pseudo-labels are wrong, they can reinforce incorrect decision boundaries and cause confirmation bias.

Vision foundation models (VFMs), such as CLIP and DINO-series visual encoders, change the role of SSL~\cite{A8,A9,A10}. 
In conventional SSL, the model must learn visual representations and class boundaries at the same time. 
In VFM-based SSL, the pretrained backbone already provides strong visual representations. 
The main challenge is therefore no longer feature learning from scratch, but reliable adaptation of a strong pretrained representation under limited labels and imperfect pseudo-label supervision. 
A common solution is to freeze the VFM backbone and update only lightweight parameter-efficient fine-tuning (PEFT) modules, such as LoRA or adapters~\cite{A11,A12}. 
This strategy reduces training cost and preserves pretrained knowledge. 
However, it also introduces a new weakness. 
Since PEFT modules have limited update capacity, their adaptation direction can be easily biased by noisy pseudo-label gradients.

Existing SSL methods usually treat pseudo-label confidence as an acceptance score~\cite{A4,A5,A6,A7}. 
High-confidence samples are selected for training, while low-confidence samples are removed or down-weighted~\cite{A4,A6,A7}. 
This binary or scalar treatment is useful, but it is not sufficient for VFM adaptation. 
Pseudo-labels do not form a simple clean-or-noisy split. 
In practice, high-confidence samples, medium-confidence samples, and low-confidence samples have different learning roles. 
High-confidence pseudo-labels are often reliable and can guide discriminative adaptation. 
Medium-confidence pseudo-labels are ambiguous, but they still contain useful alignment information near class boundaries. 
Low-confidence pseudo-labels are unstable and may contain misleading semantic signals, but simply discarding them can waste information about uncertain regions of the data distribution. 

Figure~\ref{fig:motivation_mixture} supports this observation. 
Low-confidence samples contain a high pseudo-label error rate, while high-confidence samples are much more reliable. 
However, the medium-confidence region is not useless. 
It still contains many correct pseudo-labels and occupies a non-negligible portion of unlabeled data. 
This suggests that pseudo-label supervision in SSL should be viewed as a mixture of reliability regions rather than a single uniform training signal. 
Different regions should therefore be assigned different learning roles instead of being handled by one binary selection rule.

This issue becomes more important when LoRA is used to adapt VFMs~\cite{A12}. 
LoRA learns a low-rank update on top of frozen pretrained weights~\cite{A12}. 
Its low-rank structure is efficient, but it also means that all accepted pseudo-label gradients are compressed into a small adaptation space. 
If reliable, ambiguous, and unreliable samples are all optimized by the same LoRA branch, their gradients may interfere with each other. 
Reliable samples should strengthen the correct adaptation direction. 
Ambiguous samples should improve alignment without forcing over-confident decisions. 
Unreliable samples should not dominate the update direction. 
A single LoRA module cannot explicitly separate these roles. 
As a result, standard LoRA-based SSL adaptation may either over-trust noisy pseudo-labels or under-use uncertain unlabeled samples.

\begin{figure*}[t]
    \centering
    \includegraphics[width=\linewidth]{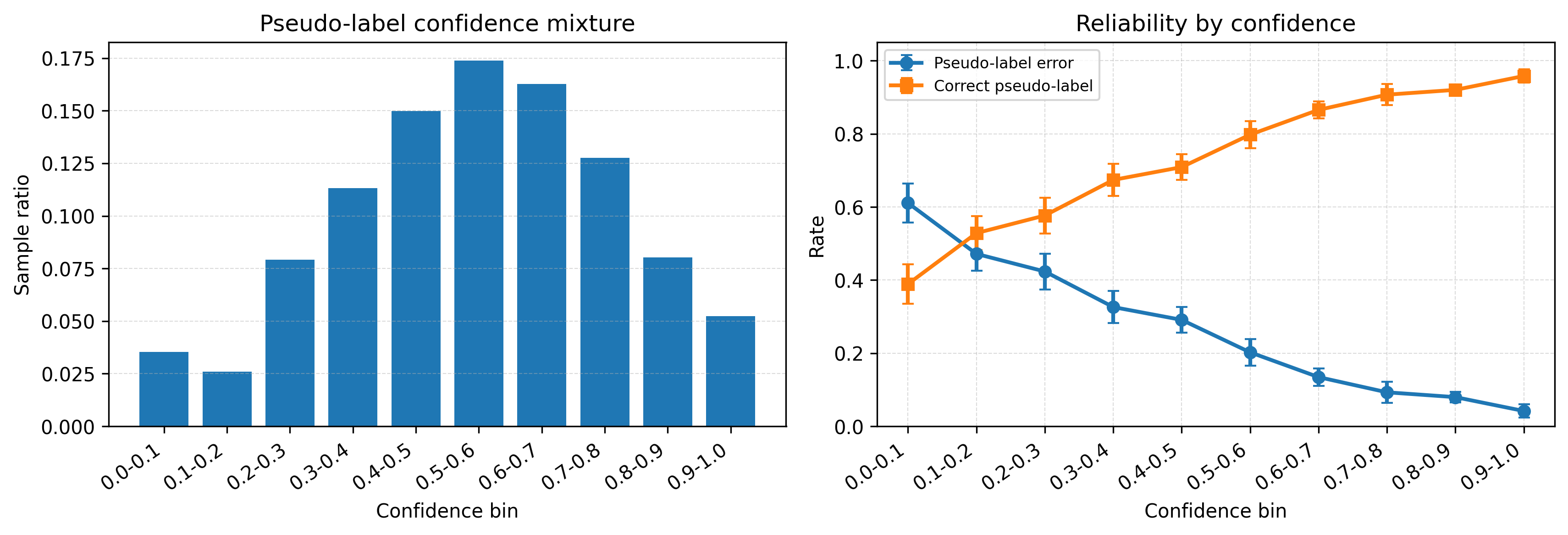}
    \vspace{-5mm}
    \caption{
    \textbf{Single LoRA suffers from mixed-gradient interference.}
    }
    \vspace{-5mm}
    \label{fig:motivation_gradient}
\end{figure*}

Figure~\ref{fig:motivation_gradient} further verifies this problem from the gradient perspective. 
A single mixed adapter shows larger deviation from the reliable update direction, which indicates that ambiguous and noisy pseudo-label gradients contaminate the same low-rank adaptation space. 
The gradient cosine curves also show that different pseudo-label regions do not provide identical optimization signals. 
The Positive and Alignment regions become increasingly compatible, while the Positive and Negative regions remain weakly conflicting. 
This supports our motivation that pseudo-label noise should not only affect the loss weight, but should also determine which adaptation pathway receives the corresponding gradient.

These observations lead to our central design principle: pseudo-label reliability should control the adaptation pathway, not only the loss weight. 
To address this problem, we propose \textbf{TriNoL}, a \textbf{Tri}ple-expert learning framework from \textbf{No}isy \textbf{L}abels for semi-supervised VFM adaptation. 
TriNoL introduces three LoRA experts to model different pseudo-label reliability regions. 
The \textbf{Positive Expert} is assigned to high-confidence pseudo-labels and learns from reliable pseudo-label supervision. 
The \textbf{Alignment Expert} is assigned to medium-confidence pseudo-labels and learns from ambiguous samples through softer alignment signals. 
The \textbf{Negative Expert} is assigned to low-confidence pseudo-labels and reduces the harmful influence of unreliable pseudo-labels. 
Importantly, the Negative Expert does not assume that all low-confidence pseudo-labels are wrong. 
Instead, it prevents uncertain pseudo-labels from being treated as clean supervision and helps protect the adaptation process from noisy gradients.

TriNoL keeps the VFM backbone frozen and only updates lightweight LoRA experts and the classifier head. 
During training, the model first predicts pseudo-labels and confidence scores from weakly augmented unlabeled samples. 
The confidence score then routes each unlabeled sample into one of three expert regions. 
High-confidence samples train the Positive Expert with hard pseudo-label supervision. 
Medium-confidence samples train the Alignment Expert with softer pseudo-label alignment. 
Low-confidence samples train the Negative Expert with a noise-aware objective that suppresses over-commitment to unreliable pseudo-labels. 
This design changes SSL adaptation from single-branch pseudo-label fitting to confidence-aware expert specialization.

The main idea behind TriNoL is that pseudo-label noise should not only affect the loss weight; it should also affect the adaptation pathway. 
Previous confidence-based SSL methods mainly decide how much each unlabeled sample contributes to the loss~\cite{A4,A5,A6,A7}. 
TriNoL instead decides which LoRA expert should absorb the corresponding adaptation signal. 
This produces a structured adaptation space: reliable gradients shape the Positive Expert, ambiguous gradients shape the Alignment Expert, and unreliable gradients are isolated by the Negative Expert. 
The three experts therefore reduce gradient interference among different pseudo-label reliability regions while keeping the parameter cost low.

Our contributions are summarized as follows. 
First, we identify a key limitation of single-adapter PEFT in semi-supervised VFM adaptation: pseudo-labels with different reliability levels are compressed into the same low-rank update space, which can cause noisy gradient interference. 
Second, we propose \textbf{TriNoL}, a confidence-aware triple LoRA expert framework that decomposes pseudo-label supervision into high-confidence, medium-confidence, and low-confidence regions. 
Instead of using one adapter for all unlabeled samples, TriNoL assigns different reliability regions to three specialized experts: the Positive Expert learns from reliable pseudo-labels, the Alignment Expert handles ambiguous samples near decision boundaries, and the Negative Expert reduces the influence of unreliable pseudo-labels without assuming that all low-confidence samples are incorrect. 
Third, we provide an efficient semi-supervised adaptation framework for VFMs, where the pretrained backbone remains frozen and only lightweight LoRA experts and the classifier head are updated. 
This makes TriNoL suitable for low-label and resource-constrained adaptation while improving robustness to noisy pseudo-label supervision.

\section{Related Work}
\label{sec:related}
\vspace{-3mm}

\textbf{Semi-supervised learning.}
Semi-supervised learning aims to improve model generalization by using both labeled and unlabeled data~\cite{A1,A2,A3,A4,A35,A36}. 
Modern SSL methods are mainly built on pseudo-labeling, self-training, and consistency regularization~\cite{A1,A2,A3,A4,A26,A37}. 
Pseudo-labeling assigns predicted labels to unlabeled samples and uses them as extra supervision~\cite{A1}, while consistency regularization encourages stable predictions under different augmentations~\cite{A2,A3,A4,A46,A47,A48,A49,A50,A51,A52,A53,A54,A55,A56,A57,A58}. 
Representative methods such as FixMatch combine weak-to-strong augmentation consistency with confidence-based pseudo-label selection~\cite{A4}. 
Later methods further improve pseudo-label selection by using curriculum learning, adaptive thresholding, soft weighting, or confidence-region bridging~\cite{A5,A6,A7,A32}. 
SSL has also been studied with stronger transformer backbones and large-scale training protocols~\cite{A17,A18,A19}. 
Although these methods are effective, they still rely on the quality of pseudo-labels. 
When pseudo-labels are noisy, the model may reinforce wrong predictions and suffer from confirmation bias~\cite{A15,A26}. 
This issue becomes more important when SSL is combined with vision foundation models, because the adaptation module is often small and sensitive to noisy gradients.

\textbf{Vision foundation models and parameter-efficient adaptation.}
Vision foundation models, such as CLIP and DINO-series encoders, provide strong pretrained visual representations~\cite{A8,A9,A10,A25,A27}. 
When they are used in downstream recognition tasks, the main problem changes from representation learning from scratch to reliable adaptation under limited labels. 
Full fine-tuning can be expensive and may damage pretrained knowledge, so parameter-efficient fine-tuning methods are commonly used~\cite{A11,A12,A15,A16,A28,A30,A31}. 
LoRA learns low-rank updates on top of frozen weights~\cite{A12}, adapter-style methods insert lightweight trainable modules into the backbone~\cite{A11,A15}, and visual prompt tuning adapts models through learnable prompts~\cite{A16}. 
These methods reduce trainable parameters and improve efficiency. 
However, their limited update capacity also makes them vulnerable to noisy pseudo-label supervision in SSL. 
If reliable and unreliable pseudo-labels are optimized through the same low-rank adapter, noisy gradients can distort the learned adaptation direction.

\textbf{Pseudo-label reliability in VFM-based SSL.}
Recent VFM-based SSL methods show that directly combining standard SSL pipelines with PEFT is not always reliable~\cite{A13,A14,A29,A33,A34}. 
A key reason is that pseudo-labels have different reliability levels. 
Existing methods usually address this issue by filtering or weighting unlabeled samples according to confidence~\cite{A4,A5,A6,A7,A32}. 
Related studies also show that calibration, pretrained knowledge, and unlabeled data quality can strongly affect SSL in the foundation model era~\cite{A14,A29,A33,A34}. 
However, these methods mainly adjust how much each sample contributes to the loss, while all pseudo-label gradients still enter the same adaptation pathway. 

Learning with noisy labels also studies how to reduce the effect of incorrect supervision. 
Representative methods use sample selection, co-training, or semi-supervised reformulation, such as Co-Teaching, JoCoR, and DivideMix~\cite{A42,A43,A44}. 
Negative and complementary learning further avoid treating noisy labels as fully correct supervision~\cite{A45}. 
These methods are related to our Negative Expert, but TriNoL focuses on semi-supervised VFM adaptation, where noisy labels are dynamically generated pseudo-labels. 
Different from prior confidence-based or noisy-label methods, TriNoL uses pseudo-label reliability as a routing signal and assigns reliable, ambiguous, and unreliable samples to different LoRA adaptation paths.

\begin{figure*}[t]
    \centering
    \includegraphics[width=\linewidth]{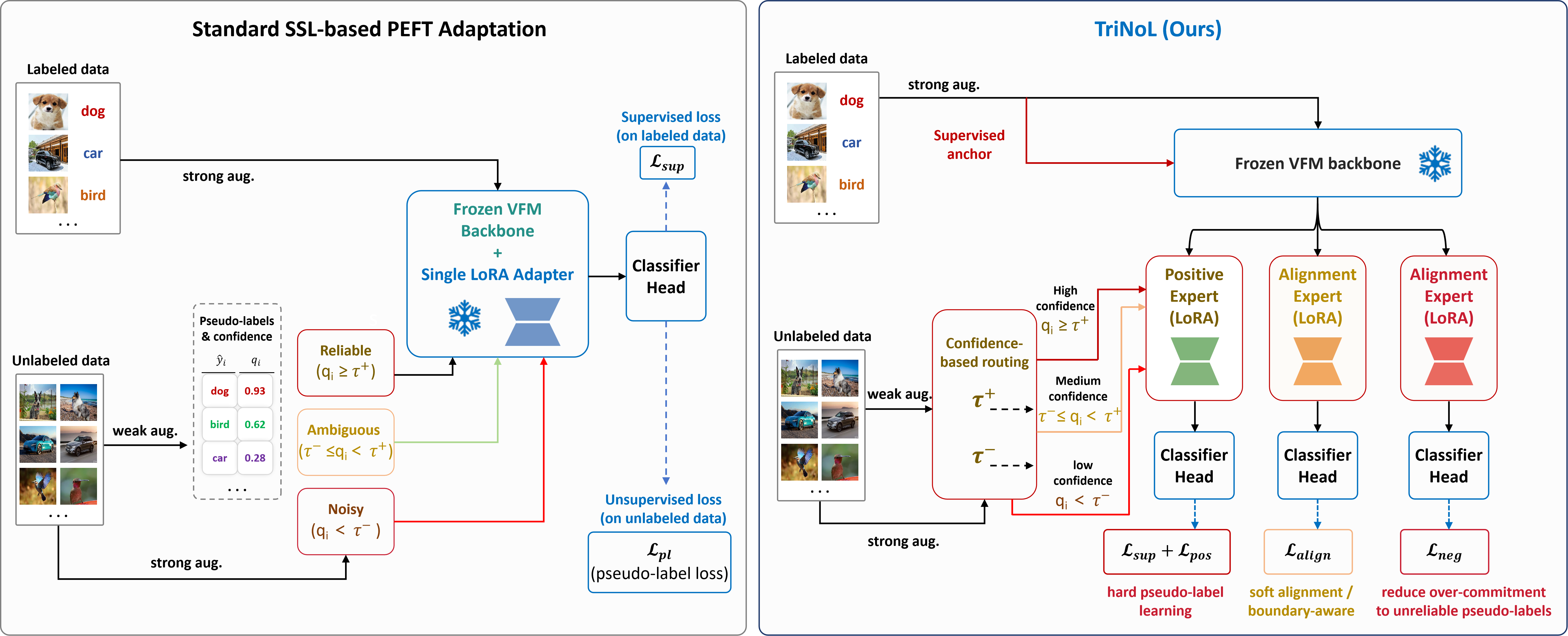}
    \vspace{-4mm}
    \caption{
    \textbf{Comparison between standard SSL-based PEFT adaptation and TriNoL.}
        Left: standard SSL-based PEFT adaptation uses a single LoRA adapter to absorb pseudo-label supervision from all unlabeled samples, including reliable, ambiguous, and noisy regions. 
    This forces mixed-quality pseudo-label gradients into the same low-rank update space. 
    Right: TriNoL decomposes unlabeled samples into three confidence regions according to pseudo-label confidence and routes them to three specialized LoRA experts. }
    \vspace{-3mm}
    \label{fig:overview}
\end{figure*}

\section{Method}
\label{sec:method}
\vspace{-3mm}

\subsection{Overview}
\label{subsec:overview}
\vspace{-3mm}

We propose \textbf{TriNoL}, a triple-expert LoRA framework for semi-supervised adaptation of vision foundation models under noisy pseudo-label supervision. 
As illustrated in Figure~\ref{fig:overview}, TriNoL differs from standard SSL-based PEFT adaptation by replacing a single LoRA adaptation path with three confidence-aware expert branches. 
TriNoL is motivated by a key observation: pseudo-labels in semi-supervised learning are not equally reliable. 
High-confidence pseudo-labels usually provide useful supervision, low-confidence pseudo-labels often contain noisy or misleading signals, and medium-confidence pseudo-labels lie in an ambiguous region where the model has not yet formed a stable decision boundary. 
However, standard parameter-efficient fine-tuning methods usually use a single adapter to absorb all pseudo-label gradients. 
This design can be fragile because a small LoRA update space may be easily affected by unreliable pseudo-labels.

To address this issue, TriNoL decomposes pseudo-label supervision into three confidence regions and assigns each region to a specialized LoRA expert. 
The \textbf{Positive Expert} learns from high-confidence pseudo-labels and captures reliable adaptation directions. 
The \textbf{Negative Expert} handles low-confidence pseudo-labels and reduces the harmful effect of unreliable supervision. 
The \textbf{Alignment Expert} focuses on medium-confidence pseudo-labels and improves the alignment of ambiguous samples near class boundaries. 
All three experts are implemented as lightweight LoRA branches, while the vision foundation model backbone remains frozen.

During training, the model first uses weakly augmented unlabeled samples to estimate pseudo-label confidence. 
According to this confidence, unlabeled samples are routed to the corresponding expert. 
High-confidence samples provide hard pseudo-label supervision to the Positive Expert. 
Medium-confidence samples provide softer alignment signals to the Alignment Expert. 
Low-confidence samples are processed by the Negative Expert to prevent noisy pseudo-labels from dominating the adaptation process. 
Labeled samples are used to maintain stable supervised learning and anchor the adaptation to ground-truth supervision.

This design turns semi-supervised VFM adaptation from a single-adapter learning problem into a confidence-aware expert learning problem. 
Instead of forcing one LoRA module to learn from mixed-quality pseudo-labels, TriNoL allows different experts to specialize in reliable, ambiguous, and unreliable supervision regions. 
As a result, the adaptation process becomes more robust to pseudo-label noise while keeping the training cost low.

\subsection{Problem Setup}
\label{subsec:problem_setup}
\vspace{-3mm}
We consider an $L$-class semi-supervised classification problem. 
Let $\mathcal{D}_l=\{(x_i,y_i)\}_{i=1}^{N_l}$ denote the labeled set, where $x_i$ is an image and $y_i \in \{1,\ldots,L\}$ is its ground-truth label. 
Let $\mathcal{D}_u=\{u_i\}_{i=1}^{N_u}$ denote the unlabeled set. 
In each training step, we sample a labeled mini-batch $\mathcal{B}_l$ and an unlabeled mini-batch $\mathcal{B}_u$. 
For each unlabeled image $u_i$, we generate a weakly augmented view $u_i^w$ and a strongly augmented view $u_i^s$.

We use a pretrained vision foundation model as the visual encoder. 
Its backbone parameters are denoted by $\psi$ and are kept frozen during training. 
A lightweight LoRA module is inserted into selected linear projections of the backbone. 
The trainable parameters include the LoRA experts and the classifier head. 
Let $C_{\omega}$ denote the classifier head with parameters $\omega$. 
For an input image $x$ and a selected expert $e$, the model prediction is written as
\begin{equation}
    z^{e}(x)=C_{\omega}\big(f_{\psi}(x;\theta^{e})\big),
\end{equation}
where $f_{\psi}(\cdot;\theta^{e})$ is the frozen VFM equipped with the LoRA expert $e$, $\theta^{e}$ denotes the expert-specific LoRA parameters, and $z^{e}(x) \in \mathbb{R}^{L}$ denotes the classification logits.

\subsection{Triple LoRA Experts}
\label{subsec:triple_lora_experts}
\vspace{-3mm}
TriNoL uses three LoRA experts to handle pseudo-labels with different confidence levels. 
The expert set is denoted as \(    \mathcal{E}=\{e^{+},e^{a},e^{-}\},\) where $e^{+}$ is the Positive Expert, $e^{a}$ is the Alignment Expert, and $e^{-}$ is the Negative Expert. 
The Positive Expert is used for high-confidence pseudo-labels. 
The Alignment Expert is used for medium-confidence pseudo-labels. 
The Negative Expert is used for low-confidence pseudo-labels.

For a frozen linear projection with weight $W \in \mathbb{R}^{d_{\mathrm{out}}\times d_{\mathrm{in}}}$, each expert learns an independent low-rank update. 
For expert $e$, the LoRA update is \(\Delta W^{e}=B^{e}A^{e},\) where $A^{e}\in\mathbb{R}^{r\times d_{\mathrm{in}}}$ and $B^{e}\in\mathbb{R}^{d_{\mathrm{out}}\times r}$ are trainable matrices, and $r$ is the LoRA rank. 
The adapted projection for expert $e$ is \(h^{e}=Wx+\alpha\Delta W^{e}x\), where $x$ is the input feature and $\alpha$ is the LoRA scaling factor. 
The frozen weight $W$ is shared by all experts, while the low-rank updates $\Delta W^{+}$, $\Delta W^{a}$, and $\Delta W^{-}$ are expert-specific.

This structure keeps the pretrained backbone fixed and only adds a small number of trainable parameters. 
More importantly, it prevents all pseudo-label gradients from being forced into the same LoRA update. 
Reliable, ambiguous, and unreliable samples can therefore shape different adaptation paths.

\subsection{Confidence-based Expert Routing}
\label{subsec:expert_routing}
\vspace{-3mm}
TriNoL uses the weakly augmented unlabeled view to estimate pseudo-label confidence. 
By default, the Positive Expert is used to produce pseudo-labels, because it is trained by labeled data and high-confidence unlabeled samples. 
For each unlabeled sample $u_i$, we compute \(p_i^w=\operatorname{softmax}\big(z^{+}(u_i^w)\big)\),

where $z^{+}(u_i^w)$ denotes the logits from the Positive Expert. 
The pseudo-label and its confidence are then given by \(    \hat{y}_i=\arg\max_{c} p_{i,c}^w, q_i=\max_{c} p_{i,c}^w \).

We use two confidence thresholds, $\tau_{+}$ and $\tau_{-}$, where $0 \leq \tau_{-}<\tau_{+}\leq 1$. 
The unlabeled samples are divided into three confidence regions:
\begin{equation}
\begin{aligned}
    \mathcal{B}_{u}^{+} &= \{u_i \in \mathcal{B}_u \mid q_i \geq \tau_{+}\}, \\
    \mathcal{B}_{u}^{a} &= \{u_i \in \mathcal{B}_u \mid \tau_{-} \leq q_i < \tau_{+}\}, \\
    \mathcal{B}_{u}^{-} &= \{u_i \in \mathcal{B}_u \mid q_i < \tau_{-}\}.
\end{aligned}
\end{equation}

Samples in $\mathcal{B}_{u}^{+}$ are routed to the Positive Expert. 
Samples in $\mathcal{B}_{u}^{a}$ are routed to the Alignment Expert. 
Samples in $\mathcal{B}_{u}^{-}$ are routed to the Negative Expert. 
This routing rule is simple and does not introduce an extra routing network. 
It uses the same confidence signal that already exists in standard SSL pipelines.

\subsection{Expert-specific Learning Objectives}
\label{subsec:expert_objectives}
\vspace{-3mm}
For labeled samples, TriNoL uses supervised learning to anchor the adaptation process. 
The labeled loss is computed with the Positive Expert:
\begin{equation}
    \mathcal{L}_{\mathrm{sup}}
    =
    \frac{1}{|\mathcal{B}_l|}
    \sum_{(x_i,y_i)\in \mathcal{B}_l}
    \operatorname{CE}
    \big(
    \operatorname{softmax}(z^{+}(x_i)), y_i
    \big),
\end{equation}
where $\operatorname{CE}(\cdot,\cdot)$ denotes the cross-entropy loss. 
This supervised term keeps the main adaptation path aligned with ground-truth labels.

For high-confidence unlabeled samples, the Positive Expert learns from hard pseudo-labels. 
The positive loss is
\begin{equation}
    \mathcal{L}_{\mathrm{pos}}
    =
    \frac{1}{|\mathcal{B}_{u}^{+}|}
    \sum_{u_i\in \mathcal{B}_{u}^{+}}
    \operatorname{CE}
    \big(
    \operatorname{softmax}(z^{+}(u_i^s)), \hat{y}_i
    \big).
\end{equation}
This term uses reliable pseudo-labels to strengthen discriminative adaptation directions.

For medium-confidence unlabeled samples, TriNoL does not force a hard pseudo-label decision. 
These samples often lie near decision boundaries, so they are better used for soft alignment. 
The Alignment Expert is trained by matching the strong-view prediction to the detached weak-view probability:
\begin{equation}
    \mathcal{L}_{\mathrm{align}}
    =
    \frac{1}{|\mathcal{B}_{u}^{a}|}
    \sum_{u_i\in \mathcal{B}_{u}^{a}}
    \operatorname{KL}
    \big(
    \operatorname{sg}(p_i^w)
    \,\Vert\,
    \operatorname{softmax}(z^{a}(u_i^s))
    \big),
\end{equation}
where $\operatorname{KL}(\cdot\Vert\cdot)$ denotes the Kullback--Leibler divergence, and $\operatorname{sg}(\cdot)$ denotes the stop-gradient operation. 
This objective preserves useful uncertainty information and avoids over-confident fitting to ambiguous samples.

For low-confidence unlabeled samples, TriNoL uses the Negative Expert to reduce the harmful effect of unreliable pseudo-labels. 
The low-confidence pseudo-label is not treated as a clean target. 
Instead, the Negative Expert is discouraged from assigning high probability to the current top pseudo-label:
\begin{equation}
    \mathcal{L}_{\mathrm{neg}}
    =
    -\frac{1}{|\mathcal{B}_{u}^{-}|}
    \sum_{u_i\in \mathcal{B}_{u}^{-}}
    \log
    \Big(
    1 -
    p_{i,\hat{y}_i}^{-}
    +
    \epsilon
    \Big),
\end{equation}
where \(p_i^{-}=\operatorname{softmax}(z^{-}(u_i^s)),\) $p_{i,\hat{y}_i}^{-}$ is the probability assigned by the Negative Expert to the weak-view top pseudo-label, and $\epsilon$ is a small constant for numerical stability. 
This loss does not assume that every low-confidence pseudo-label is wrong. 
It only prevents the model from over-committing to unreliable pseudo-labels.

The final training objective combines the supervised loss and the three expert-specific unlabeled losses:
\begin{equation}
    \mathcal{L}
    =
    \mathcal{L}_{\mathrm{sup}}
    +
    \lambda_{\mathrm{pos}}\mathcal{L}_{\mathrm{pos}}
    +
    \lambda_{\mathrm{align}}\mathcal{L}_{\mathrm{align}}
    +
    \lambda_{\mathrm{neg}}\mathcal{L}_{\mathrm{neg}},
\end{equation}
where $\lambda_{\mathrm{pos}}$, $\lambda_{\mathrm{align}}$, and $\lambda_{\mathrm{neg}}$ control the strengths of the three unlabeled objectives. 
In practice, $\lambda_{\mathrm{pos}}$ is usually larger because high-confidence pseudo-labels provide the most direct supervision. 
The alignment weight controls how much medium-confidence samples affect the decision boundary. 
The negative weight is kept moderate to avoid over-penalizing uncertain samples.

Only the LoRA experts and the classifier head are updated during training. 
The VFM backbone remains frozen. 
This preserves the pretrained representation and keeps the adaptation efficient.

The training process of TriNoL can be summarized as follows. 
First, the model predicts pseudo-labels and confidence scores from weakly augmented unlabeled samples. 
Second, unlabeled samples are divided into high-confidence, medium-confidence, and low-confidence regions. 
Third, the corresponding strongly augmented samples are routed to the Positive Expert, Alignment Expert, or Negative Expert. 
Finally, the three expert losses are optimized together with the supervised loss.

At inference time, we use the Positive Expert by default: 
\(    \hat{y}
    =
    \arg\max_{c}
    \operatorname{softmax}
    \big(
    z^{+}(x)
    \big)_c .\)

The reason is that the Positive Expert is directly trained by ground-truth labels and reliable pseudo-labels. 
The Alignment Expert and Negative Expert mainly regularize the training process by handling ambiguous and unreliable pseudo-label regions. 
Thus, TriNoL keeps inference simple while using expert specialization to improve semi-supervised training.

\section{Theoretical Analysis}
\label{sec:theory}
\vspace{-3mm}

\begingroup\small

\subsection{Single-adapter Interference under Mixed Pseudo-labels}
\label{subsec:theory_single_adapter}
\vspace{-3mm}
We first analyze why a single LoRA adapter is sensitive to mixed-quality pseudo-labels. 
For an unlabeled mini-batch, we decompose the pseudo-label gradient as
\begin{equation}
    G_{\mathrm{single}} = G_{+}+G_{a}+G_{-},
\end{equation}
where $G_{+}$ is induced by high-confidence pseudo-labels, $G_{a}$ by medium-confidence ambiguous samples, and $G_{-}$ by low-confidence noisy samples. 
The ideal reliable update should mainly follow $G_{+}$, but a single adapter is updated by the mixed gradient $G_{\mathrm{single}}$.

\begin{assumption}[Mixed pseudo-label perturbation]
\label{ass:mixed_perturbation}
The ambiguous and noisy gradients are not always aligned with the reliable gradient. 
In particular,
\begin{equation}
    \|G_{a}+G_{-}\|_2>0 .
\end{equation}
\end{assumption}

\begin{lemma}[Single-adapter gradient deviation]
\label{lem:single_deviation}
Under Assumption~\ref{ass:mixed_perturbation}, the deviation between the single-adapter update and the reliable update is
\begin{equation}
    \|G_{\mathrm{single}}-G_{+}\|_2
    =
    \|G_{a}+G_{-}\|_2
    \le
    \|G_a\|_2+\|G_-\|_2 .
\end{equation}
\end{lemma}

Lemma~\ref{lem:single_deviation} shows that a single LoRA adapter directly absorbs ambiguous and noisy pseudo-label gradients. 
Since LoRA is restricted to a low-rank update space, such mixed gradients can perturb the learned adaptation direction.

\subsection{TriNoL Reduces Reliable-update Contamination}
\label{subsec:theory_trinol_contamination}
\vspace{-3mm}
TriNoL separates different pseudo-label reliability regions into different LoRA experts. 
Let $G_{\mathrm{TriNoL}}^{+}$, $G_{\mathrm{TriNoL}}^{a}$, and $G_{\mathrm{TriNoL}}^{-}$ denote the gradients of the Positive Expert, Alignment Expert, and Negative Expert:
\begin{equation}
    G_{\mathrm{TriNoL}}^{+}=G_{+}+E_r,
    \qquad
    G_{\mathrm{TriNoL}}^{a}=G_a,
    \qquad
    G_{\mathrm{TriNoL}}^{-}=G_{-}^{\mathrm{neg}},
\end{equation}
where $E_r$ is the residual routing error into the Positive Expert, and $G_{-}^{\mathrm{neg}}$ is induced by the negative learning objective rather than hard pseudo-label fitting.

\begin{assumption}[Confidence routing quality]
\label{ass:routing_quality}
The confidence routing separates most ambiguous and noisy samples from the Positive Expert. 
There exists $0\le \rho<1$ such that
\begin{equation}
    \|E_r\|_2
    \le
    \rho \|G_a+G_-\|_2 .
\end{equation}
\end{assumption}

\begin{theorem}[Reduced reliable-update contamination]
\label{thm:reduced_contamination}
Under Assumption~\ref{ass:routing_quality}, the Positive Expert has a smaller deviation from the reliable gradient than the single adapter:
\begin{equation}
    \|G_{\mathrm{TriNoL}}^{+}-G_{+}\|_2
    \le
    \rho
    \|G_{\mathrm{single}}-G_{+}\|_2 .
\end{equation}
\end{theorem}

Theorem~\ref{thm:reduced_contamination} shows that TriNoL reduces the contamination entering the main reliable adaptation path. 
The benefit comes from routing different reliability regions into different experts, not only from changing loss weights.

\subsection{Low-rank Subspace Stability}
\label{subsec:theory_subspace}
\vspace{-3mm}
We next connect this gradient separation to LoRA's low-rank structure. 
Let $S$ denote the reliable target matrix, $A$ the ambiguous component, and $E$ the noisy component. 
A single adapter performs low-rank approximation on
\begin{equation}
    M_{\mathrm{single}}=S+A+E,
\end{equation}
while the Positive Expert performs low-rank approximation on
\begin{equation}
    M_{+}=S+E_r .
\end{equation}

\begin{assumption}[Rank-$r$ spectral gap]
\label{ass:spectral_gap}
Let $\delta_r(S)=\sigma_r(S)-\sigma_{r+1}(S)$ be the rank-$r$ singular value gap of $S$. 
Assume
\begin{equation}
    \delta_r(S)>0,
    \qquad
    \|A+E\|_{\mathrm{op}}<\frac{\delta_r(S)}{2},
    \qquad
    \|E_r\|_{\mathrm{op}}<\frac{\delta_r(S)}{2}.
\end{equation}
\end{assumption}

\begin{theorem}[Positive Expert subspace stability]
\label{thm:positive_subspace}
Under Assumption~\ref{ass:spectral_gap}, there exists an absolute constant $C>0$ such that
\begin{equation}
    \|\sin\Theta(U_r(M_{\mathrm{single}}),U_r(S))\|_{\mathrm{op}}
    \le
    C\frac{\|A+E\|_{\mathrm{op}}}{\delta_r(S)},
\end{equation}
and
\begin{equation}
    \|\sin\Theta(U_r(M_{+}),U_r(S))\|_{\mathrm{op}}
    \le
    C\frac{\|E_r\|_{\mathrm{op}}}{\delta_r(S)} .
\end{equation}
Thus, if $\|E_r\|_{\mathrm{op}}<\|A+E\|_{\mathrm{op}}$, the Positive Expert learns a low-rank subspace closer to the reliable target subspace than the single adapter.
\end{theorem}

Theorem~\ref{thm:positive_subspace} explains TriNoL from a low-rank adaptation view. 
The single adapter extracts its rank-$r$ update from a contaminated matrix, while TriNoL lets the Positive Expert approximate a cleaner reliable target.

\subsection{Roles of the Alignment and Negative Experts}
\label{subsec:theory_roles}
\vspace{-3mm}
The Alignment Expert handles medium-confidence samples without forcing hard pseudo-label decisions. 
Let $p_i^w$ be the detached weak-view probability and $p_i^a$ be the strong-view prediction from the Alignment Expert.

\begin{lemma}[Soft alignment avoids hard-label forcing]
\label{lem:soft_alignment}
For a medium-confidence sample, hard pseudo-label training gives
\begin{equation}
    \nabla_z \mathcal{L}_{\mathrm{hard}}
    =
    p_i^a-\mathrm{onehot}(\hat{y}_i),
\end{equation}
while soft alignment gives
\begin{equation}
    \nabla_z \mathcal{L}_{\mathrm{align}}
    =
    p_i^a-p_i^w .
\end{equation}
If $p_i^w$ is not one-hot, the alignment gradient preserves class uncertainty and avoids forcing all probability mass onto $\hat{y}_i$.
\end{lemma}

The Negative Expert handles low-confidence samples by reducing over-commitment to unreliable top pseudo-labels. 
For
\begin{equation}
    \mathcal{L}_{\mathrm{neg}}
    =
    -\log(1-p_{i,\hat{y}_i}^{-}+\epsilon),
\end{equation}
we have the following property.

\begin{lemma}[Negative learning suppresses unreliable top-label commitment]
\label{lem:negative_learning}
For a low-confidence sample, gradient descent on $\mathcal{L}_{\mathrm{neg}}$ decreases the probability assigned to the uncertain top pseudo-label $\hat{y}_i$. 
In contrast, hard pseudo-label cross-entropy increases the logit of $\hat{y}_i$.
\end{lemma}

Lemmas~\ref{lem:soft_alignment} and~\ref{lem:negative_learning} justify the two auxiliary experts. 
The Alignment Expert preserves uncertainty for ambiguous samples, while the Negative Expert prevents low-confidence samples from being treated as clean supervision.

\subsection{Optimization Stability}
\label{subsec:theory_stability}
\vspace{-3mm}
We finally relate the reduced pseudo-label bias to local optimization stability. 
Let $\mathcal{L}_{\mathrm{rel}}$ be the ideal reliable objective and let $\theta^*$ be its minimizer. 
At step $t$, write the stochastic update direction as
\begin{equation}
    \hat{g}_t
    =
    \nabla \mathcal{L}_{\mathrm{rel}}(\theta_t)
    +
    b_t
    +
    \xi_t,
\end{equation}
where $b_t$ is pseudo-label bias and $\xi_t$ is zero-mean stochastic noise with
\begin{equation}
    \operatorname{E}[\|\xi_t\|_2^2\mid\theta_t]\le \sigma^2 .
\end{equation}

\begin{assumption}[Local regularity]
\label{ass:local_regularity}
The reliable objective $\mathcal{L}_{\mathrm{rel}}$ is $\mu$-strongly convex and $L$-smooth in a local neighborhood of $\theta^*$. 
The pseudo-label bias is bounded as $\|b_t\|_2\le B$.
\end{assumption}

\begin{theorem}[Stability under bounded pseudo-label bias]
\label{thm:stability}
Under Assumption~\ref{ass:local_regularity}, using step size $\eta=1/L$ gives
\begin{equation}
    \limsup_{t\to\infty}
    \operatorname{E}
    \big[
    \|\theta_t-\theta^*\|_2^2
    \big]
    \le
    \frac{2B^2}{\mu^2}
    +
    \frac{2\sigma^2}{\mu L}.
\end{equation}
Therefore, if TriNoL yields a smaller bias bound than single-adapter LoRA, i.e.,
\begin{equation}
    B_{\mathrm{TriNoL}}<B_{\mathrm{single}},
\end{equation}
then TriNoL has a smaller steady-state bias radius.
\end{theorem}

Theorem~\ref{thm:stability} shows that reducing pseudo-label bias in the Positive Expert improves local optimization stability. 
Together, the Positive, Alignment, and Negative Experts reduce gradient interference while preserving the efficiency of LoRA-based VFM adaptation.

\endgroup

\section{Results and Analysis}
\label{sec:results}
\vspace{-4mm}

\noindent\textbf{Main results.}
Table~\ref{tab:main_results} reports the main comparison on CIFAR-100, FOOD-101, Semi-Aves, and ImageNet. 
TriNoL achieves the best results on CIFAR-100 N4, FOOD-101 N2/N4/N10, Semi-Aves with both in-distribution and mixed OOD unlabeled data, and ImageNet 1\%. 
These gains are most clear in low-label or noisy unlabeled settings. 
For example, on FOOD-101 N2, TriNoL improves over V-PET from $87.58\%$ to $88.42\%$, and over FineSSL from $87.27\%$ to $88.42\%$. 
On Semi-Aves with mixed in-distribution and OOD unlabeled data, TriNoL improves over FineSSL from $61.30\%$ to $61.82\%$. 
This supports our motivation that separating reliable, ambiguous, and unreliable pseudo-labels into different LoRA experts is useful when pseudo-label noise is stronger.

TriNoL does not uniformly dominate all settings. 
On CIFAR-100 N25 and N100, FineSSL and V-PET remain slightly better, and on ImageNet 10\%, FineSSL achieves the best result. 
This trend is reasonable because these settings are relatively more stable: either the label budget is larger or the pseudo-label noise has less impact on adaptation. 
In such cases, the benefit of expert separation becomes smaller. 
Overall, the results show that TriNoL is especially effective in the regimes it is designed for, namely scarce-label and noisy pseudo-label VFM adaptation, while remaining competitive in easier or more stable settings.

\begin{table*}[t]
\centering
\caption{
Performance comparisons on CIFAR-100, FOOD-101, Semi-Aves, and ImageNet. 
Mean $\pm$ std of accuracy over 3 trials are reported when available. 
For Semi-Aves, $\mathcal{D}^{u}_{in}$ denotes in-distribution unlabeled data, and $\mathcal{D}^{u}_{out}$ denotes OOD unlabeled samples. 
The best performance is highlighted in \textbf{bold}. 
All methods are evaluated under the same semi-supervised VFM adaptation setting with a frozen backbone and trainable LoRA-based adaptation modules.
}
\label{tab:main_results}
\scriptsize
\setlength{\tabcolsep}{2.0pt}
\renewcommand{\arraystretch}{0.95}
\resizebox{\textwidth}{!}{%
\begin{tabular}{lcccccccccc}
\toprule
\multirow{2}{*}{\textsc{Method}} 
& \multicolumn{3}{c}{\textbf{CIFAR-100}} 
& \multicolumn{3}{c}{\textbf{FOOD-101}} 
& \multicolumn{2}{c}{\textbf{Semi-Aves}}
& \multicolumn{2}{c}{\textbf{ImageNet}} \\
\cmidrule(lr){2-4} 
\cmidrule(lr){5-7}
\cmidrule(lr){8-9}
\cmidrule(lr){10-11}
& \textbf{N4} 
& \textbf{N25} 
& \textbf{N100} 
& \textbf{N2} 
& \textbf{N4} 
& \textbf{N10}
& $\mathcal{D}^{u}_{in}$ 
& $\mathcal{D}^{u}_{in}\cup \mathcal{D}^{u}_{out}$ 
& \textbf{1\%} 
& \textbf{10\%} \\
\midrule

\textsc{Labeled Only} 
& $64.36{\pm}0.33$ 
& $80.44{\pm}0.14$ 
& $84.62{\pm}0.29$
& $61.64{\pm}0.21$ 
& $73.92{\pm}0.22$ 
& $82.29{\pm}0.13$
& -- 
& -- 
& -- 
& -- \\

\textsc{PL}~\cite{A1}
& $66.35{\pm}0.30$ 
& $81.64{\pm}0.30$ 
& $84.84{\pm}0.20$
& $58.43{\pm}1.18$ 
& $77.78{\pm}0.93$ 
& $84.81{\pm}0.12$
& -- 
& -- 
& -- 
& -- \\

\textsc{FixMatch}~\cite{A4}
& $77.31{\pm}1.04$ 
& $84.23{\pm}0.16$ 
& $86.35{\pm}0.03$
& $75.13{\pm}1.09$ 
& $85.12{\pm}0.70$ 
& $88.91{\pm}0.02$
& $65.76{\pm}0.30$ 
& $60.46{\pm}0.16$ 
& $73.41$ 
& $78.95$ \\

\textsc{FlexMatch}~\cite{A5}
& $79.39{\pm}0.23$ 
& $83.96{\pm}0.13$ 
& $86.42{\pm}0.25$
& $81.18{\pm}1.68$ 
& $88.72{\pm}0.13$ 
& $89.43{\pm}0.11$
& $64.88{\pm}0.25$ 
& $58.38{\pm}0.11$ 
& $73.40$ 
& $78.98$ \\

\textsc{FreeMatch}~\cite{A6}
& $76.39{\pm}0.82$ 
& $83.80{\pm}0.16$ 
& $86.53{\pm}0.30$
& $74.54{\pm}4.56$ 
& $84.89{\pm}1.11$ 
& $89.15{\pm}0.24$
& $64.78{\pm}0.23$ 
& $58.35{\pm}0.38$ 
& $73.10$ 
& $78.75$ \\

\textsc{SoftMatch}~\cite{A7}
& $76.15{\pm}0.46$ 
& $84.04{\pm}0.06$ 
& $86.49{\pm}0.07$
& $76.04{\pm}5.24$ 
& $85.26{\pm}1.05$ 
& $89.20{\pm}0.08$
& $65.08{\pm}0.27$ 
& $58.24{\pm}0.52$ 
& $73.17$ 
& $78.79$ \\

\textsc{DebiasPL}~\cite{A17}
& $79.78{\pm}0.32$ 
& $84.22{\pm}0.07$ 
& $86.35{\pm}0.13$
& $86.73{\pm}0.74$ 
& $88.82{\pm}0.34$ 
& $89.52{\pm}0.17$
& $67.11{\pm}0.35$ 
& $61.07{\pm}0.37$ 
& $73.89$ 
& $79.01$ \\

\textsc{V-PET}~\cite{A31}
& $80.52{\pm}0.25$ 
& $84.65{\pm}0.05$ 
& $\textbf{86.88}{\pm}0.16$
& $87.58{\pm}0.39$ 
& $89.74{\pm}0.13$ 
& $90.05{\pm}0.14$
& -- 
& -- 
& -- 
& -- \\

\textsc{FineSSL}~\cite{A13}
& $80.66{\pm}0.24$ 
& $\textbf{84.72}{\pm}0.01$ 
& $86.86{\pm}0.18$
& $87.27{\pm}0.43$ 
& $89.49{\pm}0.14$ 
& $89.88{\pm}0.15$
& $67.51{\pm}0.11$ 
& $61.30{\pm}1.19$ 
& $74.46$ 
& $\textbf{79.42}$ \\

\midrule
\rowcolor{blue!18}
\textsc{TriNoL}
& $\textbf{80.92}{\pm}0.19$
& $84.68{\pm}0.05$
& $86.79{\pm}0.15$
& $\textbf{88.42}{\pm}0.33$
& $\textbf{90.06}{\pm}0.12$
& $\textbf{90.18}{\pm}0.11$
& $\textbf{68.04}{\pm}0.15$
& $\textbf{61.82}{\pm}0.20$
& $\textbf{74.72}$
& $79.31$ \\
\bottomrule
\end{tabular}%
}
\vspace{-3mm}
\end{table*}

\begin{table}[t]
\centering
\caption{
Efficiency comparison between Single LoRA and TriNoL.
We compare Single LoRA with rank 8, rank-matched Single LoRA with rank 24, and TriNoL with three rank-8 experts.
Params denotes trainable parameters.
LoRA cost denotes the relative LoRA forward cost normalized by Single LoRA rank 8.
}
\label{tab:efficiency}
\scriptsize
\setlength{\tabcolsep}{3.0pt}
\renewcommand{\arraystretch}{0.92}
\resizebox{\columnwidth}{!}{%
\begin{tabular}{lcccccc}
\toprule
\textsc{Method} 
& \textsc{Rank} 
& \textsc{Params} 
& \textsc{LoRA Cost} 
& \textsc{Time/Iter.} 
& \textsc{Memory} 
& \textsc{Acc.} \\
\midrule

\textsc{Single LoRA} 
& $8$ 
& $0.372$M 
& $1.0\times$ 
& $0.182$s 
& $7.8$G 
& $87.31{\pm}0.41$ \\

\textsc{Single LoRA} 
& $24$ 
& $0.962$M 
& $3.0\times$ 
& $0.205$s 
& $8.4$G 
& $87.73{\pm}0.38$ \\

\rowcolor{yellow!18}
\textsc{TriNoL} 
& $8{\times}3$ 
& $0.962$M 
& $3.0\times$ 
& $0.218$s 
& $8.7$G 
& $\textbf{88.42}{\pm}0.33$ \\

\bottomrule
\end{tabular}%
}
\vspace{-3mm}
\end{table}

\begin{figure*}[t]
    \centering
    \includegraphics[width=\linewidth]{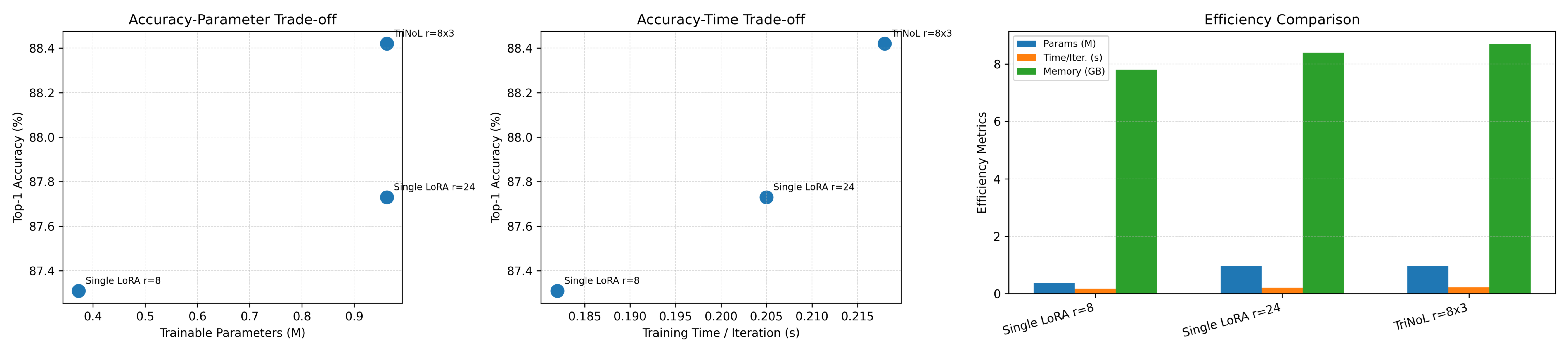}
    \vspace{-4mm}
    \caption{
    \textbf{Efficiency analysis of TriNoL.}
    We compare Single LoRA rank 8, Single LoRA rank 24, and TriNoL with three rank-8 experts.
    TriNoL has a similar trainable parameter count to Single LoRA rank 24, but achieves higher accuracy.
    This shows that the improvement mainly comes from confidence-aware expert specialization rather than merely increasing LoRA capacity.
    }
    \vspace{-3mm}
    \label{fig:efficiency}
\end{figure*}

\section{Efficiency Analysis}
\label{sec:efficiency}
\vspace{-3mm}

\noindent\textbf{Efficiency comparison.}
Table~\ref{tab:efficiency} reports the efficiency comparison among Single LoRA with rank 8, Single LoRA with rank 24, and TriNoL with three rank-8 experts. 
Single LoRA with rank 24 is included as a rank-matched baseline because it has a similar number of trainable parameters and the same relative LoRA forward cost as TriNoL. 
Compared with Single LoRA rank 8, TriNoL increases the trainable parameters from $0.372$M to $0.962$M and increases the training time per iteration from $0.182$s to $0.218$s. 
The memory cost also increases mildly from $7.8$G to $8.7$G. 
This overhead is expected because TriNoL uses three LoRA experts.

More importantly, TriNoL outperforms the rank-matched Single LoRA rank 24 baseline. 
Both methods have $0.962$M trainable parameters and the same relative LoRA cost of $3.0\times$, but TriNoL improves the accuracy from $87.73\%$ to $88.42\%$. 
This shows that the improvement does not simply come from increasing LoRA capacity. 
Instead, the gain comes from confidence-aware expert specialization, where reliable, ambiguous, and unreliable pseudo-labels are routed to different adaptation paths.

Figure~\ref{fig:efficiency} further shows the accuracy-parameter and accuracy-time trade-offs. 
TriNoL achieves the best accuracy under comparable parameter cost to Single LoRA rank 24, with only a small increase in time and memory. 
This supports our design choice: using multiple specialized low-rank experts is more effective than using a single larger LoRA adapter.

\section{Ablation Study}
\vspace{-4mm}

\begin{table*}[t]
\centering
\caption{
Component ablation of TriNoL on CIFAR-100, FOOD-101, Semi-Aves, and ImageNet.
Mean $\pm$ std of accuracy over 3 trials are reported when available.
The ablation starts from a single LoRA adapter and progressively adds the Positive, Alignment, and Negative Experts.
The best performance is highlighted in \textbf{bold}.
}
\label{tab:component_ablation}
\scriptsize
\setlength{\tabcolsep}{2.0pt}
\renewcommand{\arraystretch}{0.95}
\resizebox{\textwidth}{!}{%
\begin{tabular}{lcccccccccc}
\toprule
\multirow{2}{*}{\textsc{Setting}} 
& \multicolumn{3}{c}{\textbf{CIFAR-100}} 
& \multicolumn{3}{c}{\textbf{FOOD-101}} 
& \multicolumn{2}{c}{\textbf{Semi-Aves}}
& \multicolumn{2}{c}{\textbf{ImageNet}} \\
\cmidrule(lr){2-4} 
\cmidrule(lr){5-7}
\cmidrule(lr){8-9}
\cmidrule(lr){10-11}
& \textbf{N4} 
& \textbf{N25} 
& \textbf{N100} 
& \textbf{N2} 
& \textbf{N4} 
& \textbf{N10}
& $\mathcal{D}^{u}_{in}$ 
& $\mathcal{D}^{u}_{in}\cup \mathcal{D}^{u}_{out}$ 
& \textbf{1\%} 
& \textbf{10\%} \\
\midrule

\textsc{Single LoRA (ours)}
& $80.43{\pm}0.23$
& $84.41{\pm}0.06$
& $86.54{\pm}0.17$
& $87.31{\pm}0.41$
& $89.47{\pm}0.15$
& $89.83{\pm}0.14$
& $67.34{\pm}0.13$
& $61.18{\pm}0.32$
& $74.39$
& $79.18$ \\

\textsc{Positive Only}
& $80.57{\pm}0.21$
& $84.48{\pm}0.05$
& $86.60{\pm}0.16$
& $87.62{\pm}0.37$
& $89.66{\pm}0.13$
& $89.96{\pm}0.13$
& $67.52{\pm}0.14$
& $61.34{\pm}0.27$
& $74.47$
& $79.22$ \\

\textsc{Positive + Alignment}
& $80.76{\pm}0.20$
& $84.61{\pm}0.05$
& $86.70{\pm}0.15$
& $88.02{\pm}0.35$
& $89.88{\pm}0.13$
& $90.08{\pm}0.12$
& $67.82{\pm}0.13$
& $61.51{\pm}0.23$
& $74.61$
& $79.27$ \\

\textsc{Positive + Negative}
& $80.71{\pm}0.21$
& $84.58{\pm}0.05$
& $86.68{\pm}0.15$
& $87.94{\pm}0.36$
& $89.82{\pm}0.14$
& $90.03{\pm}0.12$
& $67.75{\pm}0.14$
& $61.62{\pm}0.22$
& $74.58$
& $79.26$ \\

\rowcolor{blue!18}
\textsc{TriNoL}
& $\textbf{80.92}{\pm}0.19$
& $\textbf{84.68}{\pm}0.05$
& $\textbf{86.79}{\pm}0.15$
& $\textbf{88.42}{\pm}0.33$
& $\textbf{90.06}{\pm}0.12$
& $\textbf{90.18}{\pm}0.11$
& $\textbf{68.04}{\pm}0.15$
& $\textbf{61.82}{\pm}0.20$
& $\textbf{74.72}$
& $\textbf{79.31}$ \\

\bottomrule
\end{tabular}%
}
\vspace{-4mm}
\end{table*}

Table~\ref{tab:component_ablation} reports the component ablation of TriNoL. 
Compared with the single LoRA variant, using only the Positive Expert already improves performance by separating high-confidence pseudo-labels from the mixed unlabeled supervision. 
Adding the Alignment Expert further improves the results, especially on FOOD-101 and Semi-Aves, which suggests that medium-confidence samples provide useful boundary information when they are trained with soft alignment instead of hard pseudo-label fitting. 
Adding the Negative Expert also brings gains, with clearer improvements under the mixed OOD setting of Semi-Aves, where low-confidence pseudo-labels are more likely to be unreliable. 
The full TriNoL model achieves the best performance across the ablation settings, showing that the three experts play complementary roles in reducing pseudo-label-induced gradient interference.

\begin{table}[t]
\centering
\caption{
Routing strategy ablation of TriNoL.
Mean $\pm$ std of accuracy over 3 trials are reported when available.
Random routing assigns unlabeled samples to experts randomly.
Shared CE uses confidence routing but trains all experts with the same hard pseudo-label loss.
TriNoL uses confidence routing with expert-specific objectives.
}
\label{tab:routing_ablation}
\scriptsize
\setlength{\tabcolsep}{3.0pt}
\renewcommand{\arraystretch}{0.95}
\resizebox{\columnwidth}{!}{%
\begin{tabular}{lccccc}
\toprule
\textsc{Routing Strategy} 
& \textsc{Expert-specific Loss}
& \textbf{CIFAR-100 N4}
& \textbf{FOOD-101 N2}
& \textbf{Semi-Aves}
& \textbf{ImageNet 1\%} \\
\midrule

\textsc{Single LoRA}
& --
& $80.43{\pm}0.23$
& $87.31{\pm}0.41$
& $61.18{\pm}0.32$
& $74.39$ \\

\textsc{Random Routing}
& \checkmark
& $80.51{\pm}0.22$
& $87.46{\pm}0.39$
& $61.25{\pm}0.29$
& $74.43$ \\

\textsc{Confidence Routing + Shared CE}
& --
& $80.69{\pm}0.21$
& $87.84{\pm}0.36$
& $61.47{\pm}0.25$
& $74.55$ \\

\rowcolor{blue!18}
\textsc{TriNoL Routing}
& \checkmark
& $\textbf{80.92}{\pm}0.19$
& $\textbf{88.42}{\pm}0.33$
& $\textbf{61.82}{\pm}0.20$
& $\textbf{74.72}$ \\

\bottomrule
\end{tabular}%
}
\vspace{-3mm}
\end{table}

\begin{figure*}[t]
    \centering
    \includegraphics[width=\linewidth]{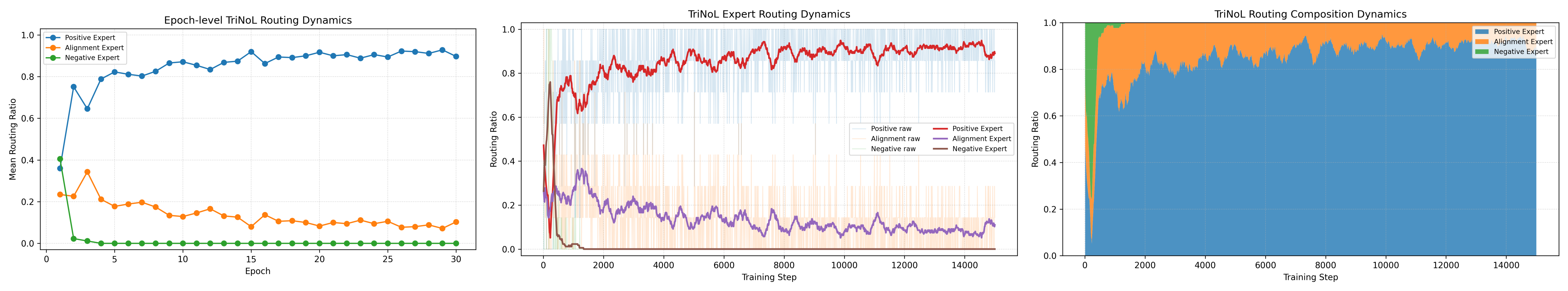}
    \vspace{-7mm}
    \caption{
    \textbf{Expert routing dynamics of TriNoL during training.}
    }
    \vspace{-3mm}
    \label{fig:expert_routing_dynamics}
\end{figure*}

\begin{figure*}[h]
    \centering
    \includegraphics[width=\linewidth]{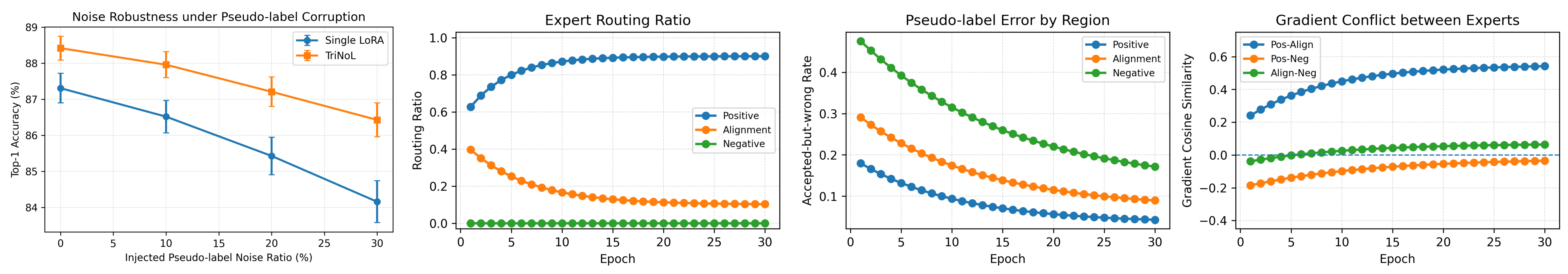}
    \vspace{-6mm}
    \caption{
    \textbf{Noise robustness and mechanism analysis of TriNoL.}
    }
    \vspace{-5mm}
    \label{fig:noise_mechanism}
\end{figure*}

\begin{table*}[t]
\centering
\caption{
Sensitivity of TriNoL to thresholds and loss weights.
}
\label{tab:sensitivity}
\tiny
\setlength{\tabcolsep}{2.0pt}
\renewcommand{\arraystretch}{0.82}
\resizebox{\columnwidth}{!}{%
\begin{tabular}{lcc|lcc}
\toprule
\multicolumn{3}{c|}{\textsc{Threshold Sensitivity}} 
& \multicolumn{3}{c}{\textsc{Loss Weight Sensitivity}} \\
\midrule
\textsc{Setting} & \textbf{Food N2} & \textbf{Semi-Aves}
& \textsc{Setting} & \textbf{Food N2} & \textbf{Semi-Aves} \\
\midrule

$(0.2,0.7)$ 
& $85.21{\pm}0.35$ 
& $61.70{\pm}0.22$
& $\lambda_a=0.5,\lambda_n=0.1$ 
& $88.28{\pm}0.34$ 
& $60.73{\pm}0.21$ \\

$(0.3,0.7)$ 
& $\textbf{88.42}{\pm}0.33$ 
& $\textbf{61.82}{\pm}0.20$
& $\lambda_a=1.0,\lambda_n=0.1$ 
& $\textbf{88.42}{\pm}0.33$ 
& $\textbf{61.82}{\pm}0.20$ \\

$(0.4,0.7)$ 
& $88.29{\pm}0.34$ 
& $61.76{\pm}0.21$
& $\lambda_a=2.0,\lambda_n=0.1$ 
& $88.31{\pm}0.36$ 
& $61.70{\pm}0.23$ \\

$(0.3,0.6)$ 
& $88.05{\pm}0.37$ 
& $61.56{\pm}0.24$
& $\lambda_a=1.0,\lambda_n=0.01$ 
& $88.18{\pm}0.37$ 
& $59.61{\pm}0.24$ \\

$(0.3,0.8)$ 
& $88.24{\pm}0.35$ 
& $61.71{\pm}0.22$
& $\lambda_a=1.0,\lambda_n=0.05$ 
& $88.34{\pm}0.34$ 
& $61.76{\pm}0.21$ \\

$(0.3,0.9)$ 
& $87.88{\pm}0.40$ 
& $61.42{\pm}0.27$
& $\lambda_a=1.0,\lambda_n=0.2$ 
& $87.11{\pm}0.39$ 
& $61.58{\pm}0.25$ \\

\bottomrule
\end{tabular}%
}
\vspace{-4mm}
\end{table*}

Table~\ref{tab:routing_ablation} studies whether the gain of TriNoL comes from the proposed confidence-aware routing or simply from using more LoRA experts. 
Random routing only brings marginal improvement over Single LoRA, which shows that adding multiple experts without reliability-aware assignment is not sufficient. 
Confidence routing with a shared hard pseudo-label loss performs better than random routing, but it is still weaker than TriNoL. 
This indicates that routing samples by confidence is useful, but each confidence region also requires a suitable learning objective. 
The full TriNoL routing achieves the best results across all representative settings, especially on FOOD-101 N2 and Semi-Aves with OOD unlabeled data. 
This supports our claim that reliable, ambiguous, and unreliable pseudo-labels should be routed to different adaptation pathways with expert-specific losses.

\noindent\textbf{Sensitivity analysis.}
Table~\ref{tab:sensitivity} studies the effect of confidence thresholds and loss weights. 
For threshold sensitivity, TriNoL performs best when $(\tau_{-},\tau_{+})=(0.3,0.7)$ on both FOOD-101 N2 and Semi-Aves. 
When $\tau_{+}$ is too low, more unreliable pseudo-labels are routed to the Positive Expert, which weakens the reliable adaptation path. 
When $\tau_{+}$ is too high, many useful pseudo-labels are moved away from the Positive Expert, reducing effective supervision. 
For loss weights, $\lambda_a=1.0$ and $\lambda_n=0.1$ give the best balance. 
A smaller alignment weight underuses medium-confidence samples, while a larger one may over-emphasize ambiguous samples. 
Similarly, a very small negative weight cannot sufficiently suppress unreliable pseudo-labels, while a large one may over-penalize uncertain samples.

\noindent\textbf{Expert routing dynamics.}
Figure~\ref{fig:expert_routing_dynamics} visualizes how unlabeled samples are routed during training. 
At early epochs, the routing distribution is less stable, and a larger portion of samples is assigned to the Alignment and Negative Experts. 
This indicates that pseudo-label reliability is still uncertain at the beginning of adaptation. 
As training progresses, the Positive Expert receives most samples, while the Alignment Expert keeps a small portion of ambiguous samples and the Negative Expert gradually becomes less active. 
This trend suggests that TriNoL does not statically split data into fixed groups. 
Instead, it dynamically transfers samples from uncertain or noisy regions to the reliable pseudo-label region as the model becomes more confident.

\noindent\textbf{Noise robustness and mechanism analysis.}
Figure~\ref{fig:noise_mechanism} further explains why TriNoL is robust to noisy pseudo-labels. 
Under injected pseudo-label corruption, TriNoL degrades more slowly than Single LoRA, showing that expert routing reduces the impact of wrong pseudo-label supervision. 
The mechanism curves also show clear separation among confidence regions. 
The Positive Expert receives samples with lower accepted-but-wrong rates, while the Negative Expert handles samples with higher error rates. 
The gradient cosine curves indicate that the Positive and Alignment Experts become increasingly aligned, whereas the Positive and Negative Experts remain weakly conflicting. 
This supports our theoretical claim that TriNoL reduces gradient interference by separating reliable, ambiguous, and unreliable pseudo-label signals into different LoRA adaptation paths.

\section{Conclusion}
\label{sec:conclusion}

In this paper, we proposed \textbf{TriNoL}, a triple-expert LoRA framework for semi-supervised vision foundation model adaptation with noisy pseudo-labels. 
TriNoL treats pseudo-labels as mixed-reliability supervision and routes unlabeled samples into three LoRA experts according to confidence. 
The Positive Expert learns from reliable pseudo-labels, the Alignment Expert handles ambiguous samples with soft alignment, and the Negative Expert suppresses over-commitment to unreliable pseudo-labels. 
This design separates different pseudo-label gradients into different adaptation paths and reduces interference in the low-rank update space. 
Experiments show that TriNoL is effective in low-label and noisy unlabeled settings while keeping the VFM backbone frozen.

\noindent\textbf{Limitations.}
TriNoL still depends on confidence-based routing, which may be affected by miscalibrated predictions or strong domain shift. 
It also uses three LoRA experts, so it has slightly higher parameter and training cost than a single rank-8 LoRA adapter. 
In addition, this work mainly studies image classification. 
Extending TriNoL to dense prediction, open-vocabulary recognition, and multimodal adaptation is left for future work.

\bibliography{egbib}

\end{document}